%% file: main.tex
\documentclass[10pt, journal]{IEEEtran}

\usepackage{amsmath,amssymb,amsfonts}
\usepackage{graphicx}
\usepackage{pifont}
\usepackage[edges]{forest}
\usepackage{tikz}
\usetikzlibrary{trees}
\usepackage[edges]{forest}
\usepackage{adjustbox}
\usepackage{amsthm}
\usepackage[figuresright]{rotating}

\usepackage{graphicx}
\graphicspath{{/}{fig/}}

\usepackage{array}
\usepackage{textcomp}
\usepackage{multirow}
\usepackage[table]{xcolor}
\usepackage{booktabs}

\usepackage{mathtools}
\usepackage{breqn}
\usepackage{float}

\usepackage{pgfplots}
\pgfplotsset{compat=1.7}
\usepgfplotslibrary{groupplots}

\usepackage{cite} 
\usepackage[utf8]{inputenc}

\usepackage{caption}
\usepackage{subcaption}

\usepackage{multirow,tabularx}
\usepackage{hyperref}
\usepackage{flushend}
\usepackage{algorithmic}
\usepackage[vlined, ruled, shortend]{algorithm2e}
\usepackage{enumitem}
\usepackage{cuted}    

\usepackage{soul}
\usepackage{wrapfig} 
\usepackage{adjustbox} 
\usepackage{makecell} 

\usepackage[usenames,dvipsnames]{xcolor}
\usepackage{tikz} \usetikzlibrary{calc, arrows.meta, intersections, patterns, positioning, shapes.misc, fadings, through,decorations.pathreplacing}

\usepackage{overpic}

\newlength\figureheight
\newlength\figurewidth
\SetAlCapNameFnt{\footnotesize}
\SetAlCapFnt{\footnotesize}

\newcommand{\cmark}{\ding{51}}%
\newcommand{\xmark}{\ding{55}}%
\title{
    Towards Embodied Agentic AI: 
    \\Review and Classification of LLM- and VLM-Driven Robot Autonomy and Interaction \\
}

\author{
    Sahar Salimpour\textsuperscript{*1}, Lei Fu\textsuperscript{*1,2}, 
    Kajetan Rachwał\textsuperscript{5,7},
    Pascal Bertrand\textsuperscript{4},
    Kevin O’Sullivan\textsuperscript{4},
    Robert Jakob\textsuperscript{4}, \\
    Farhad Keramat\textsuperscript{1}, Leonardo Militano\textsuperscript{2}, Giovanni Toffetti\textsuperscript{2}, Harry Edelman\textsuperscript{6}, Jorge Peña Queralta\textsuperscript{3,6}\\[+.42em]

     \textsuperscript{1}\href{https://www.utu.fi/en/university/faculty-of-technology/computing}{Department of Computing, University of Turku} \\[+.42em]
     \textsuperscript{2}\href{https://www.zhaw.ch/en/engineering/institutes-centres/init}{Institute of Computer Science, Zurich University of Applied Sciences} \\[+.42em]
     \textsuperscript{3}\href{https://www.zhaw.ch/en/engineering/institutes-centres/cai}{Centre for Artificial Ingelligence, Zurich University of Applied Sciences} \\[+.42em]
     \textsuperscript{4}\href{https://www.agenticsystemslab.org/}{Agentic Systems Lab, Department of Management, Technology and Economics, ETH Zürich} \\[+.42em]
     \textsuperscript{7}\href{https://ww4.mini.pw.edu.pl/}{
    Faculty of Mathematics and Information Science, Warsaw University of Technology} \\[+.42em]
     \textsuperscript{5}\href{https://robotec.ai/}{Robotec.ai} \:\: \textsuperscript{6}\href{https://binabik.ai}{Binabik.ai}\\[+.85em]
     \textsuperscript{*}\small{Authors with equal contribution. Emails: \{sahars, fakera\}@utu.fi, \{fule, milt, toff, penq\}@zhaw.ch, pbertrand@student.ethz.ch, \{kosullivan, rjakob\}@ethz.ch, kajetan.rachwal@robotec.ai, harry@binabik.ai}\vspace{-5em}

}

\begin{document}

\maketitle

\begin{strip}

\includegraphics[width=\linewidth]{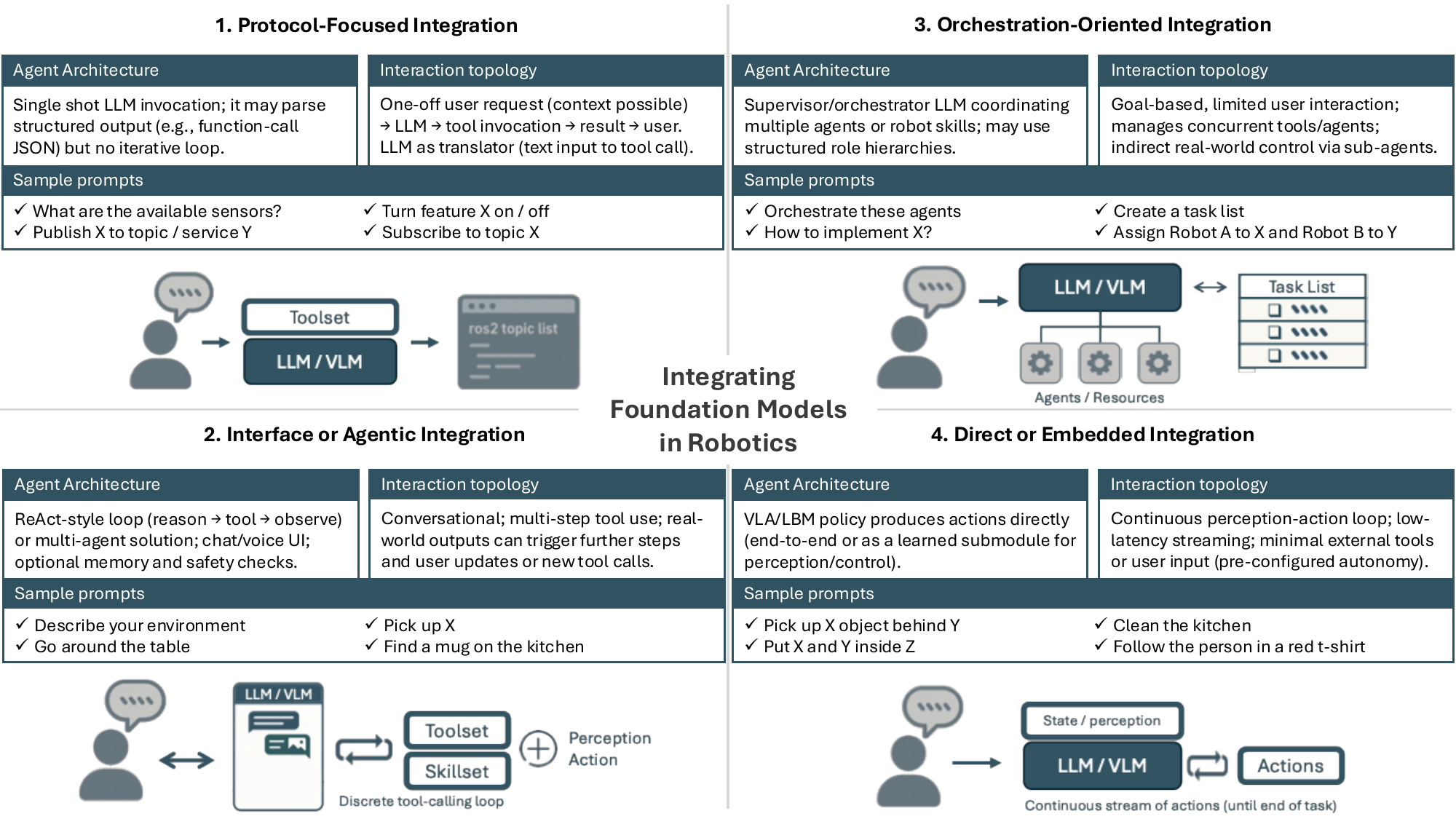}
\vspace{0.72em}
\captionof{figure}{Illustration of the main approaches to integrating large language models (LLMs), vision-language models (VLMs), and related foundation models into robotic systems. This survey
focuses on the integration perspective and the role of such models within the wider robotic system where it is deployed. Our goal is to discuss
The current state-of-the-art within the context of Embodied Agentic AI applications. The figure highlights how model integration differs in terms of \textbf{agent architecture}, \textbf{interaction topology}, and the resulting level of \textit{agency}. In \textbf{Protocol-Focused Integration}, an LLM serves as a translator between user commands and predefined APIs or protocols, typically through single-shot tool calls. \textbf{Interface or Agentic Integration} introduces ReAct-style loops and conversational interaction, allowing the agent to reason, call tools, and respond iteratively within the physical or simulated environment. \textbf{Orchestration-Oriented Integration} elevates the model to a supervisory role, coordinating multiple agents, skills, or robotic subsystems through structured workflows. Finally, \textbf{Direct or Embedded Integration} corresponds to end-to-end or model-centric policies (e.g., VLA or LBM architectures) that directly map perception to action in continuous control loops.} 
\label{fig:concept}
\end{strip}

\thispagestyle{plain}
\pagestyle{plain}

\input{sec/00_Abstract.tex}
\IEEEpeerreviewmaketitle


\input{sec/01_Intro}

\input{sec/02_Timeline}
\input{sec/03_Table}

\input{sec/04_Approaches}

\input{sec/05_Functionality}

\input{sec/06_Agent_Toolkit}
\input{sec/07_Discussion}
\input{sec/08_Conclusion}

\bibliographystyle{IEEEtran}
\bibliography{bibliography}

\end{document}

%% file: sec/00_abstract.tex
\begin{abstract}
\label{sec:abstract}
    Foundation models, including large language models (LLMs) and vision-language models (VLMs), have recently enabled novel approaches to robot autonomy and human-robot interfaces. In parallel, vision-language-action models (VLAs) or large behavior models (LBMs) are increasing the dexterity and capabilities of robotic systems. This survey paper reviews works that advance agentic applications and architectures, including initial efforts with GPT-style interfaces and more complex systems where AI agents function as coordinators, planners, perception actors, or generalist interfaces. Such agentic architectures allow robots to reason over natural language instructions, invoke APIs, plan task sequences, or assist in operations and diagnostics. In addition to peer-reviewed research, due to the fast-evolving nature of the field, we highlight and include community-driven projects, ROS packages, and industrial frameworks that show emerging trends. We propose a taxonomy for classifying model integration approaches and present a comparative analysis of the role that agents play in different solutions in today's literature.
\end{abstract}

%% file: sec/01_Intro.tex
\section{Introduction}

Large language models (LLMs) and vision-language models (VLMs) have recently enabled new modalities of interaction and reasoning in robotics. While early work focused on end-to-end pipelines, mapping raw sensory input and natural language directly to robot actions, there is growing interest in modular systems that use foundation models as high-level agents. These agents interpret user intent, generate plans, invoke robot APIs, or interface with middleware such as the Robot Operating System (ROS), without replacing the underlying robot stack.

This agentic approach offers several advantages. By integrating with existing functionality, LLMs can extend the flexibility and usability of robots without discarding tested software modules. Agents can reason over unstructured instructions, select or sequence robot capabilities, and adapt responses through interaction. Recent examples include ROS interfaces for LLMs, protocol-based control layers, and agents that coordinate multiple subsystems or external tools.

Several surveys have addressed the broader intersection of foundation models and robotics~\cite{firoozi2025foundation, urain2024deep, zeng2023large, xiao2025robot,kim2024survey}, with a focus on multimodal architectures and their components and primarily emphasizing the general use of multimodal models in robotic decision-making and high-level planning, as well as end-to-end learning frameworks for low-level control in domain-specific settings such as manipulation. However, with the very recent rise of emergence of embodied and generalist AI agents, no prior survey to date has examined the emerging design patterns in which AI agents interface with existing control software, libraries, or middleware. Moreover, many practical systems (GitHub-hosted projects, ROS packages, or startup prototypes) remain underrepresented in the literature despite their increasing real-world relevance and impact.

This survey addresses that gap. We review recent academic and community-driven work on AI agents in robotics, emphasizing architectures that position LLMs and VLMs as intelligent intermediaries rather than direct policy generators. Our contributions are twofold:
\begin{itemize}
    \item We discuss the concept of \emph{Agentic AI} in robotics and distinguish it from end-to-end learning or classical symbolic planning approaches.
    \item We propose a taxonomy across two dimensions: integration approach to foundation models (Section~\ref{sec:integration}, see also Fig.~\ref{fig:concept}), and agent role (Section~\ref{sec:roles}, see also Fig.~\ref{fig:robotic_agents_taxonomy}).
    \item We include the section (\S\ref{sec:toolkit}) presenting a practical design toolkit to guide implementation and prototyping.
    \item  Finally, the paper concludes with a detailed discussion (\S\ref{sec:openQuestions}) that highlights key open questions and outlines promising directions for future research in agentic embodied AI.
     
\end{itemize}

Together, these elements provide an alternative view, more technical and application-oriented, to the use of foundation models in robot systems.

%% file: sec/02_Timeline.tex
\input{fig/timeline}

\section{Towards Embodied Agentic AI}\label{sec:timeline}

The concept of an embodied robotic agent was introduced in the late 1980s, grounded in the idea that intelligent behavior emerges from a system’s physical embodiment and its continuous sensorimotor interaction with the environment~\cite{brooks2003robust}. This laid the foundation for early end-to-end neural models that directly mapped perception to action, enabling autonomous control without explicit symbolic reasoning~\cite{bojarski2016end, sermanet2018time, laskin2020reinforcement}. The emergence of LLMs has significantly expanded the functional scope of autonomous agents, shifting them from narrowly scoped, rule-driven systems to more flexible entities capable of generalization, contextual reasoning, and adaptive behavior~\cite{yao2023react}. At pre-ChatGPT era, Code as Policies (CaP)~\cite{liang2022code} introduced robot-executable language model generated programs (LMPs), enabling reactive and vision-based control via indirect tool calling through code generation. 

The release of OpenAI’s ChatGPT in November 2022 marked a major progression in the trajectory of embodied agentic AI, catalyzing the widespread adoption of LLMs as foundational components in robotic systems. While ROS has long been a standard framework for robot software integration, initial efforts to operationalize LLMs as autonomous agents predominantly targeted ROS-compatible platforms. By 2023, early prototypes began to explicitly integrate LLMs with robotic frameworks, mainly focusing on protocol integration and bidirectional communication interfaces between LLMs and ROS. 

Early exploration in this domain focused on natural language interfaces for ROS 2, including \textit{ros2ai}~\cite{ros2ai}, an extension to augment the ROS 2 command line interface with LLM, and ROScribe~\cite{roscribe}, a tool using LLMs to generate ROS codebases from a high-level description (e.g. a user describes a desired ROS graph (sensors, topics, and functionality) and ROScribe will output Python code for ROS nodes and launch files to implement that spec). Furthermore, the ROS-LLM framework~\cite{rosllm} was proposed to enable embodied intelligence applications. It provides a template to define your robot’s API in a config file, then handles the prompt engineering and calls to the OpenAI API, effectively acting like a customizable agent. Concurrently, Microsoft’s ChatGPT for Robotics~\cite{vemprala2024chatgpt} introduced a modular approach to LLM-based robot control and programming, defined a high-level API specific to each robot, leveraging prompt engineering and pre-defined function libraries to enable natural language control across diverse tasks and simulation environments. During this phase, most implementations adopted a chatbot-style design, deploying LLMs to interpret free-form textual inputs and generate appropriate commands for general robot integration.

Building on these advances, 2024 marked a transformative shift with the development of specialized LLM interfaces for robotics and VLA systems, unlocked unprecedented generalization and semantic reasoning capabilities in robotic control systems. These frameworks enabled agents to process open-ended natural language instructions, ground them in visual perception, and execute corresponding actions~\cite{bjorck2025gr00t}. A notable breakthrough was RT-2~\cite{zitkovich2023rt}, which introduced a novel VLA framework by co-fine-tuning pretrained VLMs on both robotic trajectory data and internet-scale vision-language tasks. $\pi$0~\cite{black2410pi0} advanced the field by introducing real-time, continuous control via flow-matched diffusion policies, unifying perception, reasoning, and motion generating in a single differentiable framework.

Simultaneously, agentic middleware frameworks such as ROSA (Robot Operating System Agent)~\cite{royce2025enabling}, RAI (Robotic AI Agent)~\cite{rachwal2025rai}, and BUMBLE~\cite{shah2024bumble} emerged, bridging the gap between LLM-based reasoning and traditional robotic middleware, in the pursuit of embodied AI. Initially open-sourced in late 2023, ROSA represents a significant advancement in AI-driven robotic control by implementing an LLM agent built on the LangChain framework and the ReAct (Reasoning and Acting) agent paradigm. Its core innovation was the abstraction of ROS operations into tool-enabled Python functions - including listing available nodes, reading sensor data, moving a joint, etc. This allowed ROSA to translate natural language commands into validated robot actions. ROSA also embedded safety mechanisms like parameter validation, constraint enforcement, and optional human approval for critical actions. The authors demonstrate ROSA across heterogeneous platforms, including JPL's NeBula-Spot quadruped and NVIDIA Isaac Sim environments. However, its architecture remains tightly coupled to ROS-based middleware, limiting interoperability with non-ROS systems and constraining deployment in broader contexts. RAI is another approach in this field as a flexible embodied multi-agent framework designed to integrate LLM reasoning with robotic systems (e.g. ROS\,2). Its architecture introduces a novel distributed paradigm where specialized LLM agents, including perception modules, task planners, motion controllers, and safety monitors, collaborate through well-defined roles to enable concurrent, real-time task execution while maintaining system safety and reliability. The framework's core components (Agents, Connectors, and Tools) provide integrated capabilies for multimodel sensing and actuation, retrieval‑augmented generation through vector stores and pre‑configured agent types (voice, conversational, state‑based). The framework demonstrates particular advancement in dynamic environments, featuring capabilities like online replanning and failure recovery. Proven on both physical (Husarion ROSBot XL) and simulated (tractor and manipulator) platforms. However, the framework is constrained by the limited spatial reasoning and unreliable self-correction of LLM-based agents, which repeatedly led to failures in understanding physical boundaries, object interactions, and task completion. BUMBLE (BUilding-wide MoBiLE Manipulation) is another recent contribution that uses a unified VLM-based framework for building-wide mobile manipulation.  It integrates open-world perception, dual-layered memory systems, and a wide range of motor skills, allowing effective operation over extended spatial and temporal horizons. Unlike frameworks such as ROSA, which rely heavily on ROS-based middleware integration, BUMBLE features tighter coupling with navigation systems through the use of predefined landmark images. While this provides efficient and precise navigation, it also introduces limitations, such as requiring prior manual collection of landmark imagery. Nevertheless, BUMBLE's combination of perception and action capabilities presents a significant advance in addressing the complexity of real-world environment.

Another notable development is the concept of an MCP (Model Context Protocol) server, driving agentic use and quick integration of new capabilities into existing tools (e.g., Anthropic's Claude or AI-driven code editors like Cursor). Their penetration in robotics is so far community-driven, with several projects openly available on GitHub. The ROS-MCP library~\cite{rosmcpserver}, for example, connects to AI assistants such as Claude and enables interaction with installed ROS\,2 applications. The capabilities include (i) launching UI applications, including the Gazebo simulator and the RViz visualizer; (ii) ROS resource management (topics, nodes); (iii) ROS system interaction (publishing to topics, calling services, send action goals); (iv) environment debugging; and (v) process management (e.g., cleaning up running ROS\,2 processes). This represents a paradigm shift from other frameworks that provide a complete application system such as ROSA and RAI, with MCP servers taking a \textit{plugin}-based approach instead, requiring standard AI assistant applications. Another example is the \textit{ros-mcp-server}~\cite{rosmcpserver}, which uses \textit{rosbridge} to connect to either ROS\,1 or ROS\,2 systems through WebSockets, and contains specific interfaces to odometry and velocity control topics. A benefit of leveraging assistants like Claude is that they already have other built-in tooling, such as web search, and the ability to run code internally in a sandboxed environment. This opens the possibility to more advanced behaviors without additional implementation. Additional implementation developed by robotmcp~\cite{robotmcp} provides a unified interface that connects LLMs with existing ROS/ROS\,2 systems without requiring source code changing, enabling true two-way integration where natural language commands are translated into ROS actions while AI can simultaneously monitor the robot states and sensor data in real time.

A final and more recent open-source framework for agentic AI in robotics is driven by the startup OpenMind~\cite{openmind}. OpenMind's OM1 introduces a modular, hardware-agnostic AI runtime intended for various robotic platforms. Compared to earlier frameworks such as ROSA and RAI, which primarily target integration with ROS-based middleware, OM1 differentiates itself with its decentralized FABRIC coordination protocol. FABRIC enables secure identity management and interoperability among heterogeneous robotic systems, diverging from the centralized orchestration common in earlier systems. While OM1 utilizes Python-based configurations and incorporates perception-action loops similar to VLA-driven systems like RT-2, it notably emphasizes decentralization. Overall, OM1 represents a step toward more collaborative and flexible robotic ecosystems.

In summary, the past three years have seen a rapid proliferation of interfaces and integration approaches for foundation models and robotics frameworks. From ROSA’s polished LangChain tools, to community-driven MCP servers, to bespoke industry solutions, the trend is clear: robots are getting a higher-level “AI agent” layer that can understand human intent and manage the robot’s own software toolkit. 

Figure~\ref{fig:timeline} illustrates in a timeline format some of the key milestones and most representative works of the past three years. The top part of the figure focuses on agentic-oriented approaches, while the bottom part lists other relevant models that have shaped the advances of the field. A more comprehensive list of works, both academic papers and community projects, is available in Table~\ref{tab:classification}. The categorization of integration approaches is discussed in detail in Section~\ref{sec:integration}. The toolset/skillset primarily reflects the mechanisms through which each system executes actions. In most works, these actions are realized through custom toolsets, referring to predefined tool classes within agentic frameworks or task-specific functions designed by developers.
In other cases—particularly for vision-based systems—the actions are driven by pretrained models or policy networks that interpret sensory inputs to perform manipulation or navigation tasks. The memory component, a fundamental aspect of agentic systems, is generally implemented in relatively common forms such as chat history or logging-based storage. In a few recent works, memory modules are extended to structured databases or multimodal memory architectures that store both textual and visual context. The world model refers not only to the representation of the physical environment, objects, and robot dynamics but also to high-level behavioral and reasoning constructs, such as system prompts that define goals, constraints, and operational rules. While the concept of a world model is closely related to the notion of adaptation, as both contribute to an agent’s ability to generalize and reason about novel situations, the world model primarily supports environmental understanding, whereas adaptability reflects the system’s capacity to update or extend its skills in response to new tasks.

\begin{table*}[t]
\centering
\caption{Comparison of Integration Approaches}
\label{tab:integration_comparison}
\renewcommand{\arraystretch}{1.2}
\begin{tabular}{@{}p{2.6cm} | p{4cm} | p{4cm} | p{5.5cm}@{}}
\toprule
\textbf{Aspect} & \textbf{Protocol-Focused Integration} & \textbf{Interface / Agentic Integration} & \textbf{Orchestration-Oriented Integration} \\
\midrule
\textbf{Main role} & LLM maps user text directly to tool/API calls & LLM acts as user-facing agent that reasons and executes actions & LLM (or small crew) coordinates multiple agents, skills, or robots \\
\textbf{Tooling focus} & Individual API or CLI calls, simple wrappers & MCP or internal tools, loops or concurrent tool calls & LangChain/LangGraph, or custom multi-agent orchestration pipelines \\
\textbf{Typical frameworks} & MCP servers, ROS CLI bindings & LangChain, LangGraph (ReAct agents), ROSA, RAI & LangGraph, custom orchestrators (e.g., AutoRT) \\
\textbf{User interaction} & Direct query–response & Conversational, human-in-the-loop reasoning & Limited direct dialogue; user gives high-level goal, orchestrator manages subtasks internally \\
\textbf{Physical interaction} & Minimal or indirect (via one API) & Yes: calls tools that affect sensors or actuators & Indirect: delegates real-world execution to sub-agents or robots \\
\textbf{Agent focus} & Translation between natural language and system commands & Reasoning + execution & Task assignment, scheduling, and monitoring \\
\textbf{Analogy} & Command translator & Assistant / operator & Project manager / mission control \\
\bottomrule
\end{tabular}
\end{table*}

%% file: fig/timeline.tex
\tikzstyle{activity} =[align=center,outer sep=1pt]

\definecolor{SlateGray}{RGB}{112,128,144}
\definecolor{SkyBlueCustom}{RGB}{135,206,235}
\definecolor{Emerald}{RGB}{80,200,120}
\definecolor{Teal}{RGB}{54,117,136}
\definecolor{OrchidCustom}{RGB}{186,85,211}
\definecolor{NavyBlue}{RGB}{0,110,184}
\definecolor{Thistle}{RGB}{216,132,183}
\definecolor{Orchid}{RGB}{175,114,176}
\definecolor{PineGreen}{RGB}{0,139,114}

\begin{figure*}[t]
\begin{tikzpicture}[very thick, black]
\small

\coordinate (O) at (-1,0); 
\coordinate (P1) at (7.5,0);
\coordinate (P3) at (12.5,0);
\coordinate (P4) at (12.75,0);
\coordinate (P5) at (13.25,0);
\coordinate (F) at (16.5,0); 

\coordinate (E1) at (3.75,0); 
\coordinate (E2) at (0.5,0); 
\coordinate (E3) at (7.5,0); 
\coordinate (E4) at (12,0); 
\coordinate (E5) at (11.25,0); 
\coordinate (E6) at (14.75,0); 
\coordinate (E7) at (15.5,0); 

\shade[left color=SlateGray, right color=SkyBlueCustom!40, opacity=0.4] (O) -- (P1) -- ($(P1)+(0,0.5)$) -- ($(O)+(0,0.5)$); 
\path [pattern color=SlateGray, pattern=north east lines, line width = 1pt, very thick] rectangle ($(O)+(0.5,0)$) -- ($(O)+(2,0)$) -- ($(O)+(2,0.5)$) -- ($(O)+(0.5,0.5)$); 
\fill[color=SkyBlueCustom!40, opacity=0.4] (P1) -- (P3) -- ($(P3)+(0,0.5)$) -- ($(P1)+(0,0.5)$); 
\shade[left color=Emerald!30, right color=white, opacity=0.4] (P3) -- (F) -- ($(F)+(0,0.5)$) -- ($(P3)+(0,0.5)$);

\shade[left color=SlateGray, right color=SkyBlueCustom!40, opacity=0.4] ($(O)+(0,0.5)$) -- ($(P1)+(0,0.5)$) -- ($(P1)+(0,1)$) -- ($(O)+(0,1)$); 
\path [pattern color=SlateGray, pattern=north east lines, line width = 1pt, very thick] rectangle ($(O)+(0.5,0.5)$) -- ($(O)+(2,0.5)$) -- ($(O)+(2,1)$) -- ($(O)+(0.5,1)$); 
\fill[color=SkyBlueCustom!40, opacity=0.4] ($(P1)+(0,0.5)$) -- ($(P4)+(0,0.5)$) -- ($(P4)+(0,1)$) -- ($(P1)+(0,1)$); 
\shade[left color=PineGreen!20, right color=white, opacity=0.4] ($(P4)+(0,0.5)$) -- ($(F)+(0,0.5)$) -- ($(F)+(0,1)$) -- ($(P4)+(0,1)$); 

\shade[left color=SlateGray, right color=SkyBlueCustom!40, opacity=0.4] ($(O)+(0,1)$) -- ($(P1)+(0,1)$) -- ($(P1)+(0,1.5)$) -- ($(O)+(0,1.5)$); 
\path [pattern color=SlateGray, pattern=north east lines, line width = 1pt, very thick] rectangle ($(O)+(0.5,1)$) -- ($(O)+(2,1)$) -- ($(O)+(2,1.5)$) -- ($(O)+(0.5,1.5)$); 
\fill[color=SkyBlueCustom!40, opacity=0.4] ($(P1)+(0,1)$) -- ($(P5)+(0,1)$) -- ($(P5)+(0,1.5)$) -- ($(P1)+(0,1.5)$); 
\shade[left color=Thistle!20, right color=white, opacity=0.4] ($(P5)+(0,1)$) -- ($(F)+(0,1)$) -- ($(F)+(0,1.5)$) -- ($(P5)+(0,1.5)$); 

\draw ($(O)+(3.5,0.8)$) node[activity,black] {Pre-ChatGPT\\Era};
\draw ($(P1)+(1.3,0.8)$) node[activity,NavyBlue] {LLM Interfaces };
\draw ($(P1)+(2.9,0.8)$) node[activity,NavyBlue] {+ VLAs};
\draw ($(P3)+(1.8,0.25)$) node[activity,SlateGray] {Open-world VLAs};
\draw ($(P4)+(1.8,0.75)$) node[activity,Teal] {World models};
\draw ($(P5)+(1.8,1.25)$) node[activity,Orchid] {General agents};
	
\draw[<-,thick,color=black] ($(E1)+(-0.15,0.1)$) -- ($(E1)+(-0.15,1.5)$) node [above=0pt,align=center,black] {Code as Policies~\cite{liang2022code}};
\draw[<-,thick,color=black] ($(E1)+(0,-0.1)$) -- ($(E1)+(0,-1)$) node [below=0pt,align=center,black] {ChatGPT Release};
\draw [decorate,decoration={brace,amplitude=6pt}]($(E2)+(-1,1.6)$) -- ($(E2)+(0.5,1.6)$) node [above=6pt,align=center,black, xshift=-22pt] {First end-to-end models \\ (e.g., self-driving~\cite{bojarski2016end})};
\draw[<-,thick,color=black] ($(E3)+(0,0.1)$) -- ($(E3)+(0,1.5)$) node [above=0pt,align=center,black] {ChatGPT for Robotics~\cite{vemprala2024chatgpt},\\Deepmind RT-2 \cite{zitkovich2023rt}};
\draw[<-,thick,color=black] ($(E4)+(0,0.1)$) -- ($(E4)+(0,1.5)$) node [above=0pt,align=center,black] {ROSA~\cite{royce2025enabling} \\ RAI~\cite{rachwal2025rai}, BUMBLE~\cite{shah2024bumble}};
\draw[<-,thick,color=black] ($(E5)+(0,-0.1)$) -- ($(E5)+(0,-1)$) node [below=0pt,align=center,black] {$\pi_0$ \cite{black2410pi0}};
\draw[<-,thick,color=black] ($(E6)+(0,-0.1)$) -- ($(E6)+(0,-1)$) node [below=0pt,align=center,black] {Gemini Robotics~\cite{team2025gemini} \\ Groot N1\cite{bjorck2025gr00t}, $\pi_{0.5}$~\cite{zhou2025vision}};
\draw[<-,thick,color=SlateGray] ($(E7)+(0,0.1)$) -- ($(E7)+(0,1.5)$) node [above=0pt,align=center,black] {OpenMind \\ OM1~\cite{openmind}};

\draw[-] (O) -- (1.5,0);           
\draw[dashed] (1.5,0) -- (4.5,0);    
\draw[->] (4.5,0) -- (F);             

\foreach \x in {0,5,9.5,14}{
    \ifdim\x pt=2.5pt\relax
      \else
        \draw (\x cm,3pt) -- (\x cm,-3pt);
      \fi
      }
\foreach \i \j in {
  0/2015,
  5/2023,
  9.5/2024,
  14/2025
}{
  \draw (\i,0) node[below=3pt] {\j};
}

\end{tikzpicture}
\caption{Progression of projects at the intersection of Generative AI and Robotics with a focus on LLMs, VLMs and VLAs.}
\label{fig:timeline}
\end{figure*}

%% file: sec/03_Table.tex
\newcolumntype{L}[1]{>{\raggedright\arraybackslash}m{#1}}  
\newcolumntype{P}[1]{>{\centering\arraybackslash}p{#1}}  
\newcolumntype{M}[1]{>{\centering\arraybackslash}m{#1}}  

\begin{sidewaystable*}
    \centering
    \setlength{\tabcolsep}{2.0\tabcolsep} %
    \renewcommand{\arraystretch}{1.5}
    \caption{Selection of most representative works spanning the time since the release of ChatGPT and the different integration approaches proposed in this survey.}
    \label{tab:classification}
    {\footnotesize
    \begin{tabular}{L{1.2cm} L{3.5cm} M{1cm} M{1.2cm} M{2.5cm} M{1.2cm} M{0.6cm} M{.8cm} M{.8cm} M{1.6cm}}
    \toprule
    \textbf{Project} & \textbf{Description} & \textbf{Year}& \textbf{Integration Approach} & \textbf{Toolset/ skillset} & \textbf{Memory}  & \textbf{Multi-modal} & \textbf{World model} & \textbf{Open-source}  & \textbf{Adaptation}\\
    \midrule
    \textbf{CaP}~\cite{liang2022code} & Robot-Centric formulation of language model programs & 2022 & Protocol &  Indirect tool through coding & Variable logging & LLM & \cmark & \cmark & N.A \\
    \textbf{ros2ai}~\cite{ros2ai} & ROS 2 command line interface extension with LLM & 2023 & Protocol &  ROS 2 CLI & \xmark & LLM & \xmark & \cmark & N.A \\
    \textbf{ROS$-$LLM} \cite{rosllm} & ROS framework for embodied intelligence applications & 2023 & Protocol & ROS actions/services or code API & Logging & LLM & \xmark & \cmark & N.A \\
    \textbf{ROScribe} \cite{roscribe} & Create ROS packages using LLMs & 2023 & Protocol/ Interface & Custom tools & - & LLM & - & \cmark & N.A \\
    \textbf{ChatGPT Robotics}~\cite{vemprala2024chatgpt} & High-level functions library for robotics application & 2023 & Protocol/ Interface & Custom tools & \xmark & LLM & \cmark & \cmark & Sys. prompt \& tools \\
    \textbf{ROSA}~\cite{royce2025enabling} & ROS Agent, interact with ROS-based robotics systems & 2024 & Interface & Custom tools & Chat history & VLM & \cmark & \cmark & Sys. prompt \& tools \\
    \textbf{RAI}~\cite{rachwal2025rai} & AI agent framework for embodied AI & 2024 & Interface/ \scriptsize{Framework} & Custom toolsets & - & VLM & \cmark & \cmark & Custom tools \\
    \textbf{Embodied Agents}~\cite{embodiedagent} & ROS 2 based framework for robot to understand models & 2024 & Interface & Custom tools & Semantic Memory & VLM & \cmark & \cmark & Sys. prompt \& tools \\
    \textbf{BUMBLE}~\cite{shah2024bumble} & Framework for open-world perception system & 2024 & Interface &  Predefined skill library &  Multimodal memory & VLM & \cmark  & \cmark & Custom skills\\
    \textbf{AutoRT}~\cite{ahn2024autort} & Models for Orchestration of Robotic Agents & 2024 & Orchestration & Pretrained RT policies & -  & VLM/ VLA  & \xmark & \xmark & Pretrained policies \\
    \textbf{ROS$-$MCP} \cite{rosmcp, rosmcpserver, robotmcp} & Converts natural language to ROS command & 2025 & Protocol &  Custom tools & - & LLM & - & \cmark & Custom tools \\
    \textbf{Gemini Robotics}~\cite{team2025gemini} & Generalist VLA model & 2025 & Embedded & Tool library & -  & VLA & - & \xmark & - \\
    \textbf{MoManip VLA}~\cite{wu2025momanipvla} & VLA for general mobile manipulation & 2025 & Embedded &  Pretrained policies & - & VLA & \xmark  & \cmark soon & Pretrained policies \\
    \textbf{ARNA}~\cite{lange2025general} & General-Purpose Robotic Navigation via LVLM & 2025 & Embedded &  Custom toolsets & Multimodal memory & VLM & \xmark  & \cmark soon & Custom tools \\
    \textbf{OpenMind OM1}~\cite{openmind} & Modular AI runtime for agents and robots & 2025 & Interface/ \scriptsize{Framework} &  Custom toolsets & Chat history & VLM & \xmark  & \cmark & Custom tools \\
    \textbf{explainable ros}~\cite{explainable} & Explanations of robot behavior using RAG method & 2024 & Protocol &  Custom toolsets & Multimodal vector database & LLM/ VLM & \xmark  & \cmark & N.A \\
    \textbf{LLama ros}~\cite{llama_ros} & Integrate llama.cpp with ROS\,2 to enable LLM inference & 2024 & Protocol/ Interface & Custom toolsets & \xmark & LLM/ VLM & \xmark  & \cmark & Custom tools \\ 
    \textbf{Stretch AI}~\cite{stretch_ai} & Orchestrates skills for language-directed mobile manipulation & 2024 & Orchestration & Custom Policies for manipulation & Dynamic memory & LLM & \xmark  & \cmark & Custom Skillsets \\
    \textbf{UROSA}~\cite{urosa} & Distributed AI Agents for Underwater Robot Autonomy & 2025 & Framework &  Custom toolsets/ \scriptsize{On the fly code generation} & Logging & VLM/ \scriptsize{LLM} & \cmark  & \cmark & Sys. prompts \\
    \bottomrule
    \end{tabular}
    }

\end{sidewaystable*}

%% file: sec/04_Approaches.tex
\section{Approaches to LLMs/VLMs/VLAs and AI Agents in Robotics}\label{sec:integration}

We propose a taxonomy to classify the work to date based on the way that large language or foundation models are integrated into robotic systems. Instead of focusing on the role, use case or type of model, our objective here is to classify the integration from a conceptual and interaction point of view. 

To that end, we propose the existence of, currently, four main directions in which foundation models are integrated into robotic systems: (i) \textit{protocol} integration, where the focus is on using the model as a \textit{translator} between protocols (e.g., user input to predefined toolset); (ii) \textit{interface} integration, where the focus is to provide an interactive approach connecting user, robotic system and the environment; (iii) \textit{orchestration-oriented} integration, where the model is in charge of managing resources, tools or subsystems; and (iv) \textit{direct} or \textit{embedded} integration, where the model acts as a perception or control policy, either end-to-end, or as a specific subsystem. These approaches are illustrated in Fig.~\ref{fig:concept}. Table~\ref{tab:integration_comparison} also provides a summary of the principal aspects and differences within this categorization. In addition, we classify a selection of representative works based on their integration approach in Table~\ref{tab:classification}.

\subsection{\textit{Protocol} Integration}

The first category, and arguably the conceptually simplest way of integrating foundation models such as LLMs into a robotic system, is to use them as \textit{translators} of user inputs.

For example, \textit{ros2ai}~\cite{ros2ai} does this by translating a generic user prompt into one specific ROS\,2 command line interface tool call. The interaction is limited, although possible, and the process is mostly \emph{unidirectional} (e.g., the user is presented with the response of the ROS command).

More recently, with the advent of the MCP protocol as the de-facto standard for tool use in LLMs/VLMs, multiple authors have proposed technical integrations~\cite{rosmcp,rosmcpserver}. The available tooling can span multiple domains: interaction with graphical interfaces (e.g., launching a simulator), the previously mentioned execution of ROS\,2 command line interface commands, or the connectivity with ROS communication paradigms (publishing/subscribing) to specific topics, calling services, or executing actions. Indeed, the MCP protocol maps well with, e.g., the service and action interfaces in ROS.

A notable related approach is code-as-policies~\cite{liang2022code}, where the LLM outputs Python code to perform the task (using a predefined API for sensorimotor actions). The code is then executed in a loop. Essentially, the LLM “programs” the robot on the fly.

\subsection{\textit{Interface} Integration}

Keeping the tool calling capabilities of a \textit{protocol} integration, but enhancing the interactiveness, we classify as \textit{interface} integrations those approaches where either (i) the output of tools (often in terms of actions in the real world) affects future commands or additional tool calls from the model; or (ii) the focus is on the user interaction, e.g., in HRI applications.

This category also goes in line of the research towards more \textit{agentic} applications, where foundation models gain further agency with respect to user commands, perform more complex tool calls, and often run in a loop with multiple iterations and multiple tool calls per iteration.

In this paradigm, user commands can stay high level, while the foundation models can decompose the command into a number of tool calls. Additionally, as in ROSA, RAI, or BUMBLE, actions or tool calls might span a certain amount of time, requiring action feedback to be given back to the user and/or the model.

The most recent and arguably complete open-source agentic system to date is OpenMind~\cite{openmind}, a commercial effort that has been demonstrated in humanoid and quadruped robots, among others, and is defining a new AI-centric ``operating system'' for intelligent robots.

\subsection{\textit{Orchestration}-Oriented Integration}

We separate \textit{interface} from \textit{orchestration} integration approaches in that the latter are focused on the problem of resource management, rather than the interaction with the user. These include mainly approaches where LLMs or VLMs act as planners or coordinators, often with other AI agents involved. 

This classification category spans both the integration approach and the role of the model, which is described in Section~\ref{sec:roles}, where we further describe planner and orchestration agent roles.

\subsection{\textit{Embedded} or \textit{Direct} Integration}

This category encompasses approaches where the LLM (or a multimodal variant) directly produces either (i) actions for the robot in an end-to-end manner, or (ii) a specific output (e.g., to be used as a perception module). These are often referred to as robotic foundation models, in their end-to-end variant. These approaches are also well covered in existing review papers~\cite{firoozi2025foundation, urain2024deep, zeng2023large, xiao2025robot}. Therefore, we only include in this survey those that are most relevant from a historical perspective, or have potential to advance towards more agentic systems. Those works are included in Table~\ref{tab:classification}.

\subsection{Agentic Frameworks}

It is important to note that more complex systems might span multiple categories. For example, a system might encompass an interface agent for user input, an internal coordinator agent, and potentially both protocol and direct agents as subsystems managed by the coordinator. However, most works to date focus on one specific integration approach, owing to the relative low maturity of the field.

In that context, we can also consider solutions such as RAI or OpenMind as frameworks, as they also provide tooling and infrastructure for multi-agent coordination and support for specialist agents. However, the current stage of development and deployment is closer to the \textit{interface} integration class.

\begin{figure*}[t]
    \centering
    \resizebox{0.99\textwidth}{10cm}{%
        \input{tables/graph} 
    }
    \caption{Classification of existing works by primary role and functionality.}
    \label{fig:robotic_agents_taxonomy}
\end{figure*}

%% file: tables/graph.tex
\begin{forest}
    for tree={
        grow'=east,
        forked edges,
        draw,
        rounded corners,
        anchor=west,
        node options={align=center},
        s sep=6mm,
        l sep=10mm,
    }
    [\adjustbox{angle=90}{\textbf{\normalsize Robotic Agents}}
        [Planner Agents, fill=red!25, text width=2cm
            [{SayCan~\cite{ahn2022can}, SELP~\cite{wu2024selp}, ConceptAgent~\cite{rivera2024conceptagent}}, fill=red!10, tier=leaf, text width=5cm
            [Focus on LLM bridging perception and action, 
            fill=red!5, rounded corners, text width=9.5cm]
            ]
            ]  
        [Orchestration Agents, fill=purple!25, text width=2cm
            [General Orchestrators, fill=purple!15, text width=3cm
                [{AutoRT~\cite{ahn2024autort},agent-orchestration~\cite{glocker2025llm}}, fill=purple!10,   tier=leaf, text width=5cm
                [{Enable scalable, environment-agnostic task assignment}, 
                    fill=purple!5, rounded corners, text width=9.5cm]]
            ]
            [Embodied Orchestrators, fill=purple!15, text width=3cm
                [{LABOR Agent~\cite{chu2024large}, SMART-LLM~\cite{kannan2024smart}, LaMMA-P~\cite{zhang2024lamma}}, fill=purple!10,   tier=leaf, text width=5cm
                [{High-level executive control across different modules and embodied agents}, 
                    fill=purple!5, rounded corners, text width=9.5cm]
                ]
            ]
        ]
        [Task-Specific Agents, fill=orange!25, text width=2cm
                [{BUMBLE~\cite{shah2024bumble}, NavGPT~\cite{zhou2024navgpt}, Cat-shaped Mug Agent~\cite{dorbala2023can},\linebreak LVLM Navigation~\cite{lange2025general}}, fill=orange!10,   tier=leaf, text width=5cm
                [{Perform reasoning and acting over navigation and manipulation tasks},
                 fill=orange!5, text width=9.5cm]
                ]
        ]
        [Model-Centric Agents, fill=blue!25, text width=2cm
                [{LEO~\cite{huang2023embodied}, RoboCat~\cite{bousmalis2023robocat}, RoboAgent~\cite{bharadhwaj2024roboagent}, Gemini Robotics~\cite{team2025gemini},$\pi_{0.5}$~\cite{zhou2025vision}}, fill=blue!10,   tier=leaf, text width=5cm
                [{End-to-end generalist agents trained on diverse data to map perception directly to action across many tasks},
                fill=blue!5,  text width=9.5cm]
                ]
        ]
        [Generalist Agents, fill=green!25, text width=2cm
            [Scalable Agentic Learning, fill=green!15, text width=3cm
                [{Voyager~\cite{wang2023voyager}, Code as Policies~\cite{liang2022code}}, fill=green!10,   tier=leaf, text width=5cm
                [{Learn, store, and reuse skills as code over time},
                fill=green!5,  text width=9.5cm]
                ]
            ]
            [LLM-Centered Skill Selectors, fill=green!15, text width=3cm
                [{ODYSSEY~\cite{liu2024odyssey}, RoboGPT~\cite{chen2025robogpt}}, fill=green!10,   tier=leaf, text width=5cm
                [{Plan subgoals based on combinations or sequences of predefined skills},fill=green!5,   text width=9.5cm]
                ]
            ]
        ]
        [Generalist Systemic Agents, fill=gray!25, text width=2cm
                [{ROSA~\cite{royce2025enabling}, RAI~\cite{rachwal2025rai}, ChatGPT for Robotics~\cite{vemprala2024chatgpt}, OpenMind~\cite{openmind}}, fill=gray!10,   tier=leaf, text width=5cm
                    [Focus on advanced tool frameworks for User-LLM interaction {(e.g., navigation, object detection)}, 
                    fill=gray!5, rounded corners, text width=9.5cm]
                ]
        ]
    ]
\end{forest}

%% file: sec/05_Functionality.tex
\section{Roles and Architectures of Robotic Agents with LLMs}\label{sec:roles}

As we surveyed recent LLM-integrated robotic systems, it became clear that many differ not in the tasks they perform, but in how they structure decision-making, control, and modularity. To capture these architectural distinctions, we group the agents according to their functional design rather than their application domain. This perspective enables a clearer comparison of how different systems reason, sequence actions, invoke tools or skills, and interface with low-level controllers. While the boundaries between categories are often blurred, since many systems integrate multiple capabilities, the classification in Figure~\ref{fig:robotic_agents_taxonomy} highlights each work’s primary role and functionality. 
\subsubsection{\textbf{Planner Agents}}
In this paradigm, an LLM plans a sequence of actions for the robot, typically by selecting or sequencing discrete skills. The LLM does not directly control actuators; instead, it outputs high-level plans that are grounded and executed by lower-level controllers. This design allows the LLM to focus on reasoning and decomposition while ensuring safety and feasibility through separate modules.

A foundational example is Google’s SayCan~\cite{ahn2022can}, where the LLM (PaLM) generates possible next actions (e.g., “pick up can” or “move to kitchen”), and a learned value function (affordance model) evaluates each option based on the robot’s current state. The most feasible action is then executed by the appropriate skill module. Similarly, SELP~\cite{wu2024selp} uses an LLM to generate symbolic task plans, followed by structured safety and efficiency filtering before execution. ConceptAgent~\cite{rivera2024conceptagent} extends this idea further by incorporating a symbolic planner and a precondition-grounding module, enabling robust task decomposition and dynamic re-planning based on environment feedback.

In contrast to orchestration agents---which continuously interact with perception, memory, and tool APIs in real-time---planning agents produce a plan upfront or iteratively refine it, with limited API interaction during execution. While orchestration agents serve as active runtime managers, planning agents focus on task structuring and leave execution monitoring to other components.

\subsubsection{\textbf{Orchestration Agents}}
As robotic systems become increasingly modular and skill-rich, a new class of agents has emerged where the LLM functions not as a planner for a single robot, but as an orchestrator that manages interactions between multiple skills, components, or even agents. These systems treat the LLM as a central controller that interprets high-level task instructions and delegates responsibility to subsystems or external tools capable of execution.

In this setting, the LLM does not execute physical actions itself, nor does it merely plan a single sequence. Instead, it performs decision-level control over which skill to invoke, which robot should act, or when to switch between behaviors. This architecture enables modularity, scalability, and adaptability, especially in multi-agent or high-skill-count environments.

For example, AutoRT~\cite{ahn2024autort} deploys an LLM to orchestrate a fleet of real-world mobile manipulators. The system uses prompt-based reasoning to map natural language instructions to specific skill invocations (e.g., grasping, opening, navigating), while integrating filtering layers to ensure safety and feasibility. Similarly, LABOR Agent~\cite{chu2024large} enables bimanual robotic manipulation by selecting among hundreds of pretrained skills. The LLM determines which skill or skill chain should be used given a complex manipulation instruction, using grounding modules to connect symbolic task descriptions to real-world motor behaviors.

In our categorization, orchestration agents primarily refer to systems that handle high-level task planning and assignment, either by coordinating multiple internal agents within a single robot~\cite{glocker2025llm}, or by distributing tasks across multiple embodied agents (i.e., multiple robots)~\cite{kannan2024smart}. While they rely on predefined capabilities, the orchestration logic remains flexible, enabling adaptive responses to novel combinations of goals, environments, or agent configurations. 

\subsubsection{\textbf{Task-Specific Agents}}
Task-specific agents are designed to solve narrowly defined problems, which used to rely on specialized training data or domain-specific policies. Recent work, however, explores how large language and vision-language models can enhance task-specific performance by enabling zero-shot reasoning or dynamic planning for particular objectives. For instance, LGX enables an embodied agent to find novel, uniquely described objects (e.g., "cat-shaped mug") using language-guided exploration and vision-language grounding without task-specific training~\cite{dorbala2023can}. Similarly, NavGPT uses explicit reasoning over visual and linguistic cues to follow navigation instructions, demonstrating that language models can guide task-specific behavior without prior tuning on navigation data~\cite{zhou2024navgpt}. ARNA leverages an LVLM to dynamically orchestrate perception and control modules for general-purpose navigation, but it can also be framed as a task-specific agent when executing goal-directed exploration in unseen environments~\cite{lange2025general}. Similarly, BUMBLE demonstrates a highly capable agent for manipulation and navigation by integrating an LLM with a world model and structured memory, enabling it to follow user instructions across tasks.

\subsubsection{\textbf{Model-Centric Agents}}
Unlike traditional VLA models that often require separate components for perception, reasoning, and action, each trained or tuned independently, model-centric agents adopt a unified architecture where a single model is responsible for grounding multimodal inputs (e.g., images, language, proprioception) and directly producing action outputs. Examples include LEO~\cite{huang2023embodied}, which leverages a decoder-only large language model to integrate 2D egocentric vision, 3D point clouds, and text for both instruction following and physical interaction in 3D environments; RoboCat~\cite{bousmalis2023robocat}, which uses a goal-conditioned decision transformer to generalize across robotic embodiments and tasks through large-scale training and self-improvement; and RoboAgent~\cite{bharadhwaj2024roboagent}, which achieves high data efficiency and broad task generalization through semantic augmentation and action chunking within a unified policy model.

\subsubsection{\textbf{Generalist Agents}}
This category of agents reflects a shift toward foundation-style architectures for robotics, where a centralized reasoning model (often an LLM or a multimodal transformer) interfaces flexibly with grounded, executable components. Although each system may differ in how tools are created, selected, or executed, they share a common philosophy: general-purpose reasoning layered on top of task-specialized capabilities.

Generalist agents are typically built for multi-task, multi-domain operation. Their architectures are modular by design: the high-level model interprets goals, decomposes them into subtasks, and delegates execution to a set of low-level modules, which are usually predefined skills, learned action models, or custom tools. These execution modules may be static (pretrained) or dynamic (created and expanded during deployment).

For example, Voyager~\cite{wang2023voyager} is an open-ended generalist agent that autonomously explores new tasks in Minecraft by generating its own tools (as Python functions), evaluating them, and storing them for future use---effectively building a lifelong, self-curated skill library. Similarly, Code as Policies uses an LLM to generate executable Python policies that map observations directly to robot actions. ODYSSEY~\cite{liu2024odyssey} likewise operates in an open-world setting, using an LLM to reason about tasks and select from a rich library of skills grounded in the environment. In the real world, RoboGPT~\cite{chen2025robogpt} demonstrates generalist behavior by interpreting language instructions and invoking appropriate pretrained skills for manipulation and navigation, without needing task-specific retraining.

What distinguishes these agents is not just their task breadth, but their ability to generalize across tasks through modular reasoning and flexible skill integration---an architectural pattern increasingly central to the future of scalable robotic intelligence.

\subsubsection{\textbf{Generalist Systemic Agents}}
Generalist Systemic Agents focus on building reusable, modular frameworks that simplify the development and orchestration of LLM-based robotic systems. Rather than targeting specific tasks, these approaches—such as ROSA, and RAI emphasize system-level design, where perception, reasoning, and action modules are cleanly separated and easily composed. Many of these agents are built upon general-purpose LLM frameworks like LangChain, enabling flexible memory, tool integration, and conversational interfaces. Notably, this category closely overlaps with the \textit{Agentic Frameworks} paradigm discussed in Section~\ref{sec:integration}, as these systems rely on structured interfaces through which the LLM can invoke robotic capabilities.

%% file: sec/06_Agent_Toolkit.tex
\section{Toolkits for Agentic Physical AI}\label{sec:toolkit}

Early agent systems in artificial intelligence, such as symbolic planners, belief–desire–intention (BDI) architectures, and the SOAR cognitive framework, laid the groundwork for connecting reasoning with action. The advent of deep learning shifted the focus to data-driven methods, and more recently, LLMs have enabled agents to integrate reasoning, perception, and action in naturalistic ways. Toolkits for \emph{agentic physical AI} now aim to close the loop between high-level reasoning and embodied control, making it possible to deploy agents in robotics and other physical domains. This section surveys the main components of this emerging ecosystem.

\subsection{Large Language Models and Providers}
State-of-the-art LLMs form the backbone of agentic systems. Closed-source providers such as OpenAI (GPT-4o), Anthropic (Claude~3.5), and Google DeepMind (Gemini~1.5) offer advanced reasoning and multimodal capabilities, along with integrated tool use. Open-source efforts include Meta (Llama~3.1), Mistral AI (Mixtral MoE), and xAI (Grok). These models differ in openness, context length, and function-calling capabilities, but all increasingly support structured integration with external systems.

\subsection{Tool Calling}

While impressive, LLMs originally operated only through isolated textual interfaces limited to their training data, without the ability to directly access real-time information, external tools, or APIs. To bridge this gap, one of the first major steps was function calling, introduced by OpenAI, where models were fine-tuned to recognize when a function should be invoked and respond with structured JSON that matched the function’s signature~\cite{buadicua2025contemporary}. Over time, this approach evolved into what is now broadly known as tool calling (renamed from function calling), which is supported by most modern LLMs. However, implementations differ across vendors in how JSON structures are defined and handled directly by LLMs. The Model Context Protocol (MCP) builds on this idea by introducing an open, interoperable standard that enables consistent and reliable communication between clients (LLM-powered applications) and servers (e.g., tools). Unlike vendor-specific tool calling, MCP ensures that hosts can access tools, resources, and real-time context updates in a standardized way without constant custom integration work.

In the MCP, there are three key components: the MCP Server, MCP Client, and MCP Host. The MCP Server is a program that provides context in the form of tools (callable functions or actions), resources (structured data or knowledge), and prompts (predefined templates). A server can define multiple of these and communicate them using the JSON-RPC 2.0 protocol, either locally through stdio transport or remotely through streaming HTTP. The MCP Client is a protocol-level component that connects to exactly one server and is responsible for handling this communication, passing the context back to the host. Finally, the MCP Host is any AI application that integrates one or more MCP Clients to extend its capabilities; examples include Claude Desktop, Cursor, VS Code with MCP integrations, and agentic frameworks such as LangChain.

\subsection{Frameworks for Agentic Systems}
Several frameworks provide abstractions for building complex, multi-step, and multi-agent workflows. These frameworks vary in their focus (retrieval, orchestration, or collaboration) but share the goal of simplifying agent construction and deployment. In this section, we explore key aspects and highlight the use cases of several prominent frameworks, with more detailed attention to the LangChain and on top of that LangGraph framework, which has gained notable popularity in recent agentic robotics systems.

\subsubsection{LangGraph/LangChain} 
LangGraph is a graph-based framework (nodes and edges) built on top of LangChain. To situate it briefly, LangChain is a modular framework that provides structured components for building tool-integrated and memory-aware agents based on LLMs. While LangChain focuses on modularity, LangGraph introduces a state-machine paradigm and supports conditional, non-linear workflows such as loops and retrievals. This makes it particularly suitable for building robust agent pipelines and even coordinating multiple interacting agents.  

In LangGraph, each \textit{node} represents a discrete piece of logic—this can be a call to an LLM, a function for data processing, or even another agent—while the \textit{edges} define the control flow, including conditional branching and iterative loops. Several features are especially relevant for designing agentic physical AI systems:
\begin{itemize}
    \item \textit{Pre-built implementations:} LangGraph provides ready-to-use templates for common agent patterns. Examples include:
    \begin{itemize}
        \item \textit{Single-tool calling ReAct:} Implements the ReAct pattern, combining reasoning traces with tool use in a structured way.  
        \item \textit{Supervisor multi-agent system:} Individual agents are coordinated by a central supervisor agent, which controls all communication flow and task delegation. The supervisor decides which agent to invoke based on the current context and task requirements.  
        \item \textit{Swarm multi-agent system:} Agents dynamically hand off control to one another depending on their specializations. The system tracks which agent was last active so that subsequent interactions resume with the same agent, ensuring conversational and task continuity.  
    \end{itemize}
    \item \textit{MCP adapters:} The framework integrates with the Model Context Protocol (MCP), making it easier to connect external tools and services in a standardized way.  
    \item \textit{Human-in-the-Loop capabilities:} Developers can insert checkpoints where human users review or approve agent outputs before execution continues, ensuring safety and reliability.  
    \item \textit{Memory:} LangGraph natively supports short- and long-term memory, enabling agents to retain context across steps or sessions, which is critical for persistent autonomy in physical environments.  
\end{itemize}

\subsubsection{LlamaIndex}
LlamaIndex specializes in RAG, providing structured ingestion and query pipelines over heterogeneous data sources. For agentic physical AI, this allows agents to ground decisions in sensor logs, spatial maps, or multimodal datasets. It is frequently paired with orchestration frameworks such as LangGraph for more complex workflows.

\subsubsection{CrewAI}
CrewAI focuses on multi-agent collaboration, where agents are assigned roles and tasks within a coordinated ``crew.'' This makes it suitable for collaborative problem solving, distributed control, and simulation.

\subsubsection{AutoGen}
AutoGen (Microsoft) emphasizes conversational multi-agent systems, supporting both AI--AI and AI--human collaboration. Its modular design makes it extensible to embodied AI scenarios, where agents must interact with humans and physical systems in tandem.

\subsection{Domain-specific considerations}

Robotic data and control pipelines present unique challenges for agentic integration due to their inherent multimodality and temporal diversity. Typical robotic systems combine heterogeneous data streams (vision, laser ranging, tactile sensing, audio, joint encoders, or radio telemetry) each with distinct payload sizes, update frequencies, and latency tolerances. High-bandwidth modalities such as camera or depth feeds require continuous streaming and compression strategies, while others, like status or event messages, demand reliability and strict quality-of-service guarantees. Designing an agentic layer that can flexibly operate across these heterogeneous data rates and protocols remains nontrivial. Agents must reason over asynchronous information and decide which data to process, summarize, or ignore in real time, often under constraints of compute, bandwidth, and safety.

Another domain-specific constraint lies in embodiment variability. A general-purpose agent should be capable of interfacing with multiple robotic platforms; even if constrained to mobile agents, at least: mobile bases, mobile manipulators, drones, or legged robots (e.g., quadrupeds or humanoids). Each of these embodiments comes with different control interfaces and sensing configurations. Middlewares such as ROS or ROS\,2 already provide abstraction layers, from abstracting hardware drivers with topics, services, and actions, to providing out-of-the-box manipulation (e.g., moveit) or navigation (e.g., nav2) capabilities. However, the degree to which an LLM- or VLM-driven agent can remain at a high level of reasoning depends on the embodiment. While protocol-based and interface-level integrations can remain platform-agnostic, models that generate low-level actions or policies (e.g., VLA or RL-based systems) still face challenges in transferring across morphologies. Generalizing such embodied control requires richer world models and modular skill representations that can decouple reasoning from actuation.

Finally, many real-world robotic tasks are inherently long-horizon and involve unpredictable interaction with the physical environment. Unlike static API calls, physical actions introduce variable execution times, uncertainties, and potential failure modes. Tasks such as “navigate to gate A42” in an airport, or “deliver this object to the suite in the top floor”, involve navigation through dynamic, human-centric spaces, possibly requiring waiting, replanning, or the use of elevators and constrained corridors. For an agentic autonomy solution, reasoning about such extended temporal dependencies (potentially spanning minutes or hours) requires continuous situational awareness, progress monitoring, and adaptive recovery strategies. These factors underscore that agentic systems in robotics must not only manage data and tools, but also reason about time, uncertainty, and embodiment, bridging symbolic decision-making with the continuous realities of physical interaction.

%% file: sec/07_Discussion.tex
\section{Discussion}\label{sec:openQuestions}

Agentic Embodied AI is a rapidly developing field where research is arguably in its infancy. While embedded policies such as VLAs or BLMs are relatively mature, the approaches to integrating foundation models into existing or new autonomy stacks are still underexplored. Below, we outline some of the key open questions and promising directions for future work.

\subsection{AI Agents, AI Workflows and Agentic AI}

The deployment of LLMs across systems with differing levels of autonomy, decision logic, and task scope necessitates a conceptual distinction between AI Workflows, AI Agents, and Agentic AI Systems. These categories differ primarily in how control, reasoning, and coordination are distributed between humans, models, and the surrounding infrastructure.

AI Workflows operate along predefined, human-engineered paths in which the sequence of tool use and model calls is explicitly programmed. Here, the LLM acts as a component within a deterministic pipeline, executing a bounded subtask such as summarization. This approach prioritizes predictability, control, and auditability, making it well-suited for regulated environments or applications requiring reproducible outputs~\cite{anthropic2024building}.

In contrast, AI Agents are systems that delegate control of task execution to the model itself. Rather than following a fixed sequence, the LLM dynamically decides which actions to take: planning, executing, and iterating toward a defined goal. The agent perceives its environment, evaluates progress, and selects subsequent actions such as invoking tools. This autonomy enables agents to tackle open-ended problems that cannot be easily pre-specified in a workflow. However, this flexibility comes at the expense of predictability, interpretability, and verifiability ~\cite{anthropic2024building, sapkota2505ai}.

Finally, Agentic AI Systems represent a higher level of organization, in which multiple autonomous agents collaborate, coordinate, and share memory to achieve complex or long-horizon objectives. Rather than focusing on individual problem-solving capabilities, the emphasis shifts to the emergent behaviors that arise from interaction, negotiation, and collective reasoning among agents. This conceptual progression, from isolated task execution toward emergent multi-agent organization, is increasingly discussed in the context of robotics and autonomous systems~\cite{sapkota2505ai}.

\subsection{Embodiment}

Embodiment can be understood as the capacity of a cyber-physical system to process information about its physical component.
It can be viewed as a spectrum: some systems possess limited awareness of their physical properties, while others exhibit more advanced self-modeling.
At the simplest level, this involves reasoning about the system’s own physical capabilities: its actuator modalities or “skills,” as outlined in the Internet of Skills~\cite{guzman2022ios}.
On the other hand, it has been argued that true embodiment requires a more comprehensive internal model.
Duffy and Joue~\cite{duffy2000embodiment} introduced the notion of strong embodiment, describing agents that not only understand their actuator modalities but also use this understanding to explore their environment, reason about it, and infer its physical laws.

Therefore, embodiment can be understood not only as an agent’s capacity to reason about its own body, but also about the environment in which it operates.
It is thus better viewed not as an intrinsic property of the agent, but as a trait emerging from the dynamic coupling between the agent and its environment.
Consequently, an agent’s embodiment may be context-dependent: a robot designed to perceive and act within a structured warehouse may lose embodiment when deployed in an open field.

For this reason, in this work we adopt the broadest possible definition of embodiment when conducting our survey.
All systems considered possess some capacity to model and reason about their physical component within their operational context.
While this capacity differs in form and degree (ranging from direct sensorimotor feedback to abstract physical modeling) each system can nonetheless be regarded as embodied.
This diversity shows that embodiment is not a binary property but a spectrum shaped by how agents process information within, and about their environments.

\subsection{Applications to Multi-Robot/agent Systems}

In this work, we focus on embodied agentic AI solutions, in contrast to other surveys that explicitly categorize into single-agent and multi-agent approaches~\cite{feng2025multi}. The term multi-agent does not necessarily imply multi-robot; its interpretation varies across frameworks. For instance, in systems such as RAI~\cite{rachwal2025rai}, a single robot may host multiple agents, whereas in orchestration architectures, one agent may control multiple robots. Furthermore, many frameworks adopt a one-to-one mapping between robots and agents~\cite{kannan2024smart}. Nonetheless, understanding the distribution of agents among robots remains an important aspect, as it reintroduces the debate between centralized and decentralized coordination architectures.

A key challenge is the development of general multi-robot/agent embodied AI frameworks. While recent progress has largely targeted constrained tasks, current methods struggle with scalability, heterogeneous agent teams, and variable objectives. Extending the principles of single-agent generalist models to multi-agent settings introduces difficulties such as non-stationarity, intricate interactions, and multiple equilibria. Unlike the closed and static scenarios often assumed in research, real-world deployments involve dynamic, uncertain, and adversarial conditions with noisy observations, shifting tasks, and fluid team compositions. Achieving robust performance in such environments requires resilient policies, continual learning, and context-aware reasoning to adapt rapidly to novel agents and evolving tasks.

Finally, applications across domains highlight both the promise and the challenges of embodied multi-robot/agent AI. Robotics for manufacturing, logistics, and autonomous driving demand scalable coordination and adaptability; education and healthcare require socially appropriate, ethical, and privacy-preserving interactions; and defense applications necessitate robustness under adversarial conditions with human oversight. Beyond these, smart cities, simulations, and energy systems call for efficient, trustworthy, and safe agent collectives. Addressing these diverse needs will require progress in multimodal perception, collaborative and social learning, and ethical governance, ensuring that multi-robot embodied AI evolves into a reliable, interpretable, and impactful technology for society.  

\subsection{Ethical consideration}

Ethical issues arise when considering the deployment of embodied agents with humans: the possibility of physical harm, misinformation, or privacy violations.
These are compounded by social threats posed by such systems, such as the rapidly changing labor market, and lack of sufficient policies regarding responsibility~\cite{perlo2025embodied}.
They are said to fail to address the division of obligations between manufacturers of such systems, model providers, and organizations deploying them in the real world.
Addressing these distinctions is crucial for responsible development and application of embodied AI systems.
However, it comes with a complex set of challenges.

It could be argued that introducing legislation forcing agentic AI system providers to comply with regulation could lead to a decrease in competition by making it harder for new companies to join the market~\cite{Mueller2021AIActCost}~\cite{UKGov2023AIGovernanceImpact}.
Finding a balance between appropriate regulation, while not stifling innovation, is crucial for successful and ethical deployment of AI-driven cyber-physical systems.

\subsection{AI Agents in the Edge-Cloud Continuum}

The increasing distribution of computational intelligence across robots, edge nodes, and cloud infrastructures introduces new challenges for embodied agentic AI. Robotic agents often operate in dynamic conditions where compute, storage, and connectivity must be balanced across heterogeneous devices and administrative domains. Agentic systems must therefore reason about where and how to execute perception, reasoning, and planning processes, dynamically orchestrating components in response to context and performance requirements. Distributed and federated approaches have begun to address this by enabling semantic task decomposition, runtime component migration, and adaptive resource selection across the compute continuum~\cite{Acharya2025, Filinis2026, Baccour2022}. These methods emphasize low-latency execution for perception at the edge while offloading heavy reasoning or model adaptation to the cloud, supporting a resilient and scalable paradigm for embodied autonomy. As such, the edge–cloud continuum becomes not merely an infrastructure layer but an integral part of the embodied agent’s cognitive architecture.

\subsection{Memory}

Long-term and contextually grounded memory remains one of the least mature capabilities in current agentic architectures. Although tool-based integration servers, such as Bagel~\cite{bagel} and rosbag MCP, provide standardized access to past sensor data and execution traces~\cite{ros2mcp2025}, most existing agents still operate over transient, session-bound context windows. The lack of persistent, semantically indexed memory prevents agents from maintaining situational continuity, sharing experience between tasks, or reasoning over past states in real time. Emerging frameworks explore structured and distributed memory abstractions to support multi-session persistence and retrieval across agents~\cite{Ehtesham2025, Mei2025}, yet real-time synchronization and consistency remain open problems. Overcoming these limitations will require live, queryable memory layers capable of bridging symbolic reasoning with embodied experience—enabling agents to continuously learn, adapt, and recontextualize actions across time and embodiment.

\subsection{Continuous data streams}

Integrating data streams with LLMs, VLMs, and Embodied AI systems remains an active area of research.

In robotics, a central challenge lies in handling continuous data streams from sensors such as cameras, LiDAR, and radar. These streams provide real-time sensor input, but conflict with the discrete mode of operation of most LLMs and VLMs.
Continuous data must therefore be transformed into suitable representations.
For example, a 10 Hz video feed cannot be directly used by an LLM which takes more than 0.1 seconds to process an image.
That would lead to desynchronization and potential information loss.

Recent work explores strategies for selecting and summarizing data streams.
For example, Hu~\cite{hu2025mllmbasedvideoframe} introduced a method using a multimodal LLM for the selection of frames from a video.
Similarly, Lin~\cite{lin2025captionskeyframeskeyscoremultimodal} proposed a keyframe identification model that uses captioning and scoring to preserve context.
Such approaches highlight promising directions for aligning continuous perception with discrete reasoning.
However, integrating streaming data into embodied AI systems remains an open challenge.

\subsection{World Models}

The concept of \emph{world models} has recently gained significant attention in the field of robotics and artificial intelligence. World models are internal representations that an agent (e.g., a robot) constructs to reason about its environment, predict the consequences of its actions, and plan its behavior.
One of the pioneering works on world models is the paper by Ha and Schmidhuber\cite{world_models_2018}, which proposed a framework for building compact, compressed representations of the world using deep neural networks.
The authors demonstrated the effectiveness of world models in various reinforcement learning tasks, where the agent learns to navigate and complete tasks in simulated environments.
Following the work of Ha and Schmidhuber, several researchers have explored different aspects of world models and their applications in robotics. For a recent survey please see \cite{Kawaharazuka_2024}.
Recent works are for instance \cite{hafner2024masteringdiversedomainsworld}, proposing DreamerV3, a general algorithm that outperforms specialized methods across over 150 diverse tasks, or \cite{hansen2024tdmpc2scalablerobustworld}, introducing TD-MPC2, a model-based RL algorithm designed for learning generalist world models on large non-curated datasets composed of multiple
task domains, embodiments, and action spaces. The authors of \cite{zhou2024dinowmworldmodelspretrained} present DINO-WM, a method for training visual models by using pretrained DINOv2 embeddings of image frames: once trained, given a target observation, one can directly optimize agent behavior by planning through DINO-WM using model-predictive control (MPC).

World models can be valuable tools to enhance the robot's understanding of the environment, more specifically to gauge the consequences/effects of its actions on the environment. Coupled with an exploratory strategy (e.g., an agentic loop) to assess what possible actions could be undertaken to perform a specific task and their predicted effects, the robot can choose to enact the most promising strategies from the world model in the real world,  leading to safer and more robust task execution. While a world model is learnt and possibly updated from "experience", another similar exploration approach could be implemented by using simulation.

\subsection{Evaluation Metrics}

Evaluation metrics for embodied agentic systems can be defined along the Perceive–Reason–Act paradigm.
Task performance is often measured through success rate and completion time, particularly in multi-stage navigation and manipulation tasks.
The efficiency of tool and function definition significantly influences these outcomes, as clear and adaptable tool descriptions improve task execution regardless of the underlying tool-use capabilities of the foundation models.
A critical aspect of evaluating physical agentic systems is the level of autonomy, defined as the ability of a system to plan, decide, and act without constant human intervention.
Many existing frameworks demonstrate a medium level of autonomy, as they rely on predefined commands, functions, or tool sets.
Achieving higher autonomy requires agents to generate their own sub-goals and corresponding actions.
In some settings, especially those involving limited human interaction, this trait may be desirable.
However, evaluating systems with higher autonomy is much more difficult.

Systems with a low degree of autonomy, i.e., those which utilize advanced tooling, can be often evaluated through monitoring the order and arguments of the tools.
The higher the degree of autonomy, the need arises to evaluate the system through its impact on the environment in which it operates.
This can lead to some undesirable behaviors, such as excessive movement, being missed during the evaluation.
For example, in a highly autonomous agent responsible for its own path planning, simply checking whether it reached the goal may overlook inefficient or excessive movement.
That issue is not present in low-autonomy systems using predefined navigation tools.

Adaptability is another key factor, reflecting the agent’s ability to generalize across unseen tasks, environments, or robotic embodiments with minimal additional data or modification.
Memory consistency and recallability assess how effectively an agent stores, retains, and reuses contextually relevant information, serving as a foundation for stable, context-aware behavior.
Other factors, such as safety, transparency, are also essential in evaluating embodied AI agents~\cite{raptis2025agentic} and have been extensively discussed in recent surveys on agentic AI~\cite{bandi2025rise}.

%% file: sec/08_Conclusion.tex
\section{Conclusion}

This survey reviewed the emerging field of agentic AI in robotics, where large language and vision-language models function as intelligent intermediaries between users and robotic systems. Unlike end-to-end learning approaches, agentic systems preserve the structure of traditional robotics pipelines, using LLMs to translate goals, generate plans, and invoke APIs or tools based on available capabilities.

We organized the current landscape into two key taxonomic dimensions: model integration patterns, and agent role. Across all dimensions, we included both academic literature and practical systems from the open-source community, or industrial applications. Most of the literature today can be classified under one specific area, e.g., with a focus on either \textit{protocol}, \textit{interface}, \textit{orchestration}, or \textit{embedded} integration of foundation models. However, there are already a number of works that propose more complex system that span multiple areas. Similarly, from the perspective of agent roles, model-centric agents operate as either orchestration or generalist agents. Therefore, we see significant value in combining different approaches and subsystems to achieve higher degrees of intelligence.

Embodied or Physical Agentic AI for robotics is still in an early but rapidly evolving stage. Future work will need to address challenges related to grounding, memory, safety, deployment efficiency, and evaluation. As modular agent frameworks continue to improve, we expect increased adoption across industrial and academic robotics, making intelligent interaction with complex systems more accessible, interpretable, and adaptive.

%% file: bibliography.bib
@article{Acharya2025,
  author    = {D. B. Acharya and P. Gupta and L. He and J. Morales},
  title     = {Agentic AI: Autonomous Intelligence for Complex Goals},
  journal   = {IEEE Access},
  year      = {2025},
  volume    = {13},
  pages     = {21548--21563},
  doi       = {10.1109/ACCESS.2025.10849561}
}

@article{Filinis2026,
  author    = {N. Filinis and K. Mei and A. Ehtesham and F. Strati},
  title     = {Edge AI-driven Robotic Applications in the Computing Continuum},
  journal   = {IEEE Internet of Things Magazine},
  note      = {submitted},
  year      = {2026}
}

@article{Baccour2022,
  author    = {E. Baccour and M. Chen and S. K. Das},
  title     = {Pervasive AI for IoT Applications: A Survey on Resource-Efficient Distributed Artificial Intelligence},
  journal   = {IEEE Communications Surveys \& Tutorials},
  year      = {2022},
  volume    = {24},
  number    = {4},
  pages     = {2366--2418},
  doi       = {10.1109/COMST.2022.3189539}
}

@article{Ehtesham2025,
  author    = {A. Ehtesham and R. Sapkota and D. B. Acharya},
  title     = {A Survey of Agent Interoperability Protocols: Model Context Protocol (MCP), Agent Communication Protocol (ACP), Agent-to-Agent Protocol (A2A), and Agent Network Protocol (ANP)},
  journal   = {arXiv preprint},
  eprint    = {2504.16736},
  year      = {2025},
  url       = {https://arxiv.org/abs/2504.16736}
}

@inproceedings{Mei2025,
  author    = {K. Mei and J. Li and X. Zhang},
  title     = {AIOS: LLM Agent Operating System},
  booktitle = {Proceedings of the Conference on Language Modeling (COLM)},
  year      = {2025},
  address   = {Vienna, Austria}
}

@article{ros2mcp2025,
  author    = {Lei Fu and Sahar Salimpour and Leonardo Militano and Harry Edelman and Jorge {Pe{\~n}a Queralta} and Giovanni Tofetti},
  title     = {ROSBag MCP Server Analyzing Robot Data with LLMs for Agentic Embodied AI Applications},
  journal   = {arXiv preprint},
  eprint    = {2503.09842},
  year      = {2025},
  url       = {https://arxiv.org/abs/2503.09842}
}

@misc{hansen2024tdmpc2scalablerobustworld,
      title={TD-MPC2: Scalable, Robust World Models for Continuous Control}, 
      author={Nicklas Hansen and Hao Su and Xiaolong Wang},
      year={2024},
      eprint={2310.16828},
      archivePrefix={arXiv},
      primaryClass={cs.LG},
      url={https://arxiv.org/abs/2310.16828}, 
}

@misc{zhou2024dinowmworldmodelspretrained,
      title={DINO-WM: World Models on Pre-trained Visual Features enable Zero-shot Planning}, 
      author={Gaoyue Zhou and Hengkai Pan and Yann LeCun and Lerrel Pinto},
      year={2024},
      eprint={2411.04983},
      archivePrefix={arXiv},
      primaryClass={cs.RO},
      url={https://arxiv.org/abs/2411.04983}, 
}

@article{Kawaharazuka_2024,
   title={Real-world robot applications of foundation models: a review},
   volume={38},
   ISSN={1568-5535},
   url={http://dx.doi.org/10.1080/01691864.2024.2408593},
   DOI={10.1080/01691864.2024.2408593},
   number={18},
   journal={Advanced Robotics},
   publisher={Informa UK Limited},
   author={Kawaharazuka, Kento and Matsushima, Tatsuya and Gambardella, Andrew and Guo, Jiaxian and Paxton, Chris and Zeng, Andy},
   year={2024},
   month=sep, pages={1232–1254} }

@misc{hafner2024masteringdiversedomainsworld,
      title={Mastering Diverse Domains through World Models}, 
      author={Danijar Hafner and Jurgis Pasukonis and Jimmy Ba and Timothy Lillicrap},
      year={2024},
      eprint={2301.04104},
      archivePrefix={arXiv},
      primaryClass={cs.AI},
      url={https://arxiv.org/abs/2301.04104}, 
}

@article{world_models_2018,
  doi = {10.5281/ZENODO.1207631}, 
  url = {https://zenodo.org/record/1207631},  
  author = {Ha, David and Schmidhuber, Jürgen},  
  title = {World Models},  
  publisher = {Zenodo},  
  year = {2018},  
  copyright = {Creative Commons Attribution 4.0}
}

@article{zhou2025vision,
  title={Vision-Language-Action Model with Open-World Embodied Reasoning from Pretrained Knowledge},
  author={Zhou, Zhongyi and Zhu, Yichen and Wen, Junjie and Shen, Chaomin and Xu, Yi},
  journal={arXiv preprint arXiv:2505.21906},
  year={2025}
}

@article{bojarski2016end,
  title={End to end learning for self-driving cars},
  author={Bojarski, Mariusz and Del Testa, Davide and Dworakowski, Daniel and Firner, Bernhard and Flepp, Beat and Goyal, Prasoon and Jackel, Lawrence D and Monfort, Mathew and Muller, Urs and Zhang, Jiakai and others},
  journal={arXiv preprint arXiv:1604.07316},
  year={2016}
}

@misc{ros2ai,
  howpublished = {ros2ai: \url{https://github.com/fujitatomoya/ros2ai}}
}

@misc{rosllm,
  howpublished = {ROS-LLM: \url{https://github.com/Auromix/ROS-LLM}}
}

@misc{roscribe,
  howpublished = {ROSCribe: \url{https://github.com/RoboCoachTechnologies/ROScribe}}
}

@misc{rosmcp,
  howpublished = {ROS-MCP: \url{https://github.com/Yutarop/ros-mcp}}
}

@misc{rosmcpserver,
  howpublished = {ROS-MCP-Server: \url{https://github.com/lpigeon/ros-mcp-server}}
}

@misc{embodiedagent,
  howpublished = {EmbodiedAgents: \url{https://automatika-robotics.github.io/embodied-agents/intro.html}}
}

@misc{bagel,
  howpublished = {Bagel: \url{https://github.com/Extelligence-ai/bagel}}
}

@misc{openmind,
  howpublished = {OpenMind: \url{https://docs.openmind.org/developing/0_introduction}}
}

@misc{robotmcp,
  howpublished = {ROS-MCP-Server by robotmcp: \url{https://github.com/robotmcp/ros-mcp-server}}
}

@article{lange2025general,
  title={General-Purpose Robotic Navigation via LVLM-Orchestrated Perception, Reasoning, and Acting},
  author={Lange, Bernard and Yildiz, Anil and Arief, Mansur and Khattak, Shehryar and Kochenderfer, Mykel and Georgakis, Georgios},
  journal={arXiv preprint arXiv:2506.17462},
  year={2025}
}

@article{firoozi2025foundation,
  title={Foundation models in robotics: Applications, challenges, and the future},
  author={Firoozi, Roya and Tucker, Johnathan and Tian, Stephen and Majumdar, Anirudha and Sun, Jiankai and Liu, Weiyu and Zhu, Yuke and Song, Shuran and Kapoor, Ashish and Hausman, Karol and others},
  journal={The International Journal of Robotics Research},
  volume={44},
  number={5},
  pages={701--739},
  year={2025},
  publisher={SAGE Publications Sage UK: London, England}
}

@article{urain2024deep,
  title={Deep generative models in robotics: A survey on learning from multimodal demonstrations},
  author={Urain, Julen and Mandlekar, Ajay and Du, Yilun and Shafiullah, Mahi and Xu, Danfei and Fragkiadaki, Katerina and Chalvatzaki, Georgia and Peters, Jan},
  journal={arXiv preprint arXiv:2408.04380},
  year={2024}
}

@article{zeng2023large,
  title={Large language models for robotics: A survey},
  author={Zeng, Fanlong and Gan, Wensheng and Wang, Yongheng and Liu, Ning and Yu, Philip S},
  journal={arXiv preprint arXiv:2311.07226},
  year={2023}
}

@article{xiao2025robot,
  title={Robot learning in the era of foundation models: A survey},
  author={Xiao, Xuan and Liu, Jiahang and Wang, Zhipeng and Zhou, Yanmin and Qi, Yong and Jiang, Shuo and He, Bin and Cheng, Qian},
  journal={Neurocomputing},
  pages={129963},
  year={2025},
  publisher={Elsevier}
}

@article{brooks2003robust,
  title={A robust layered control system for a mobile robot},
  author={Rodney Brooks},
  journal={IEEE journal on robotics and automation},
  volume={2},
  number={1},
  pages={14--23},
  year={2003},
  publisher={IEEE}
}

@inproceedings{sermanet2018time,
  title={Time-contrastive networks: Self-supervised learning from video},
  author={Sermanet, Pierre and Lynch, Corey and Chebotar, Yevgen and Hsu, Jasmine and Jang, Eric and Schaal, Stefan and Levine, Sergey and Brain, Google},
  booktitle={2018 IEEE international conference on robotics and automation (ICRA)},
  pages={1134--1141},
  year={2018},
  organization={IEEE}
}

@article{laskin2020reinforcement,
  title={Reinforcement learning with augmented data},
  author={Laskin, Misha and Lee, Kimin and Stooke, Adam and Pinto, Lerrel and Abbeel, Pieter and Srinivas, Aravind},
  journal={Advances in neural information processing systems},
  volume={33},
  pages={19884--19895},
  year={2020}
}

@inproceedings{yao2023react,
  title={React: Synergizing reasoning and acting in language models},
  author={Yao, Shunyu and Zhao, Jeffrey and Yu, Dian and Du, Nan and Shafran, Izhak and Narasimhan, Karthik and Cao, Yuan},
  booktitle={International Conference on Learning Representations (ICLR)},
  year={2023}
}

@article{ahn2022can,
  title={Do as i can, not as i say: Grounding language in robotic affordances},
  author={Ahn, Michael and Brohan, Anthony and Brown, Noah and Chebotar, Yevgen and Cortes, Omar and David, Byron and Finn, Chelsea and Fu, Chuyuan and Gopalakrishnan, Keerthana and Hausman, Karol and others},
  journal={arXiv preprint arXiv:2204.01691},
  year={2022}
}

@article{wu2024selp,
  title={SELP: Generating safe and efficient task plans for robot agents with large language models},
  author={Wu, Yi and Xiong, Zikang and Hu, Yiran and Iyengar, Shreyash S and Jiang, Nan and Bera, Aniket and Tan, Lin and Jagannathan, Suresh},
  journal={arXiv preprint arXiv:2409.19471},
  year={2024}
}

@article{rivera2024conceptagent,
  title={Conceptagent: Llm-driven precondition grounding and tree search for robust task planning and execution},
  author={Rivera, Corban and Byrd, Grayson and Paul, William and Feldman, Tyler and Booker, Meghan and Holmes, Emma and Handelman, David and Kemp, Bethany and Badger, Andrew and Schmidt, Aurora and others},
  journal={arXiv preprint arXiv:2410.06108},
  year={2024}
}

@article{ahn2024autort,
  title={Autort: Embodied foundation models for large scale orchestration of robotic agents},
  author={Ahn, Michael and Dwibedi, Debidatta and Finn, Chelsea and Arenas, Montse Gonzalez and Gopalakrishnan, Keerthana and Hausman, Karol and Ichter, Brian and Irpan, Alex and Joshi, Nikhil and Julian, Ryan and others},
  journal={arXiv preprint arXiv:2401.12963},
  year={2024}
}

@inproceedings{chu2024large,
  title={Large language models for orchestrating bimanual robots},
  author={Chu, Kun and Zhao, Xufeng and Weber, Cornelius and Li, Mengdi and Lu, Wenhao and Wermter, Stefan},
  booktitle={2024 IEEE-RAS 23rd International Conference on Humanoid Robots (Humanoids)},
  pages={328--334},
  year={2024},
  organization={IEEE}
}

@inproceedings{zitkovich2023rt,
  title={Rt-2: Vision-language-action models transfer web knowledge to robotic control},
  author={Zitkovich, Brianna and Yu, Tianhe and Xu, Sichun and Xu, Peng and Xiao, Ted and Xia, Fei and Wu, Jialin and Wohlhart, Paul and Welker, Stefan and Wahid, Ayzaan and others},
  booktitle={Conference on Robot Learning},
  pages={2165--2183},
  year={2023},
  organization={PMLR}
}

@inproceedings{royce2025enabling,
  title={Enabling novel mission operations and interactions with rosa: The robot operating system agent},
  author={Royce, Rob and Kaufmann, Marcel and Becktor, Jonathan and Moon, Sangwoo and Carpenter, Kalind and Pak, Kai and Towler, Amanda and Thakker, Rohan and Khattak, Shehryar},
  booktitle={2025 IEEE Aerospace Conference},
  pages={1--16},
  year={2025},
  organization={IEEE}
}

@article{rachwal2025rai,
  title={RAI: Flexible Agent Framework for Embodied AI},
  author={Rachwal, Kajetan and Majek, Maciej and Boczek, Bartlomiej and Dabrowski, Kacper and Liberadzki, Pawel and Dabrowski, Adam and Ganzha, Maria},
  journal={arXiv preprint arXiv:2505.07532},
  year={2025}
}

@article{vemprala2024chatgpt,
  title={Chatgpt for robotics: Design principles and model abilities},
  author={Vemprala, Sai H and Bonatti, Rogerio and Bucker, Arthur and Kapoor, Ashish},
  journal={Ieee Access},
  volume={12},
  pages={55682--55696},
  year={2024},
  publisher={IEEE}
}

@article{shah2024bumble,
  title={Bumble: Unifying reasoning and acting with vision-language models for building-wide mobile manipulation},
  author={Shah, Rutav and Yu, Albert and Zhu, Yifeng and Zhu, Yuke and Martin-Martin, Roberto},
  journal={arXiv preprint arXiv:2410.06237},
  year={2024}
}

@article{dorbala2023can,
  title={Can an embodied agent find your “cat-shaped mug”? llm-based zero-shot object navigation},
  author={Dorbala, Vishnu Sashank and Mullen, James F and Manocha, Dinesh},
  journal={IEEE Robotics and Automation Letters},
  volume={9},
  number={5},
  pages={4083--4090},
  year={2023},
  publisher={IEEE}
}

@article{bousmalis2023robocat,
  title={Robocat: A self-improving generalist agent for robotic manipulation},
  author={Bousmalis, Konstantinos and Vezzani, Giulia and Rao, Dushyant and Devin, Coline and Lee, Alex X and Bauz{\'a}, Maria and Davchev, Todor and Zhou, Yuxiang and Gupta, Agrim and Raju, Akhil and others},
  journal={arXiv preprint arXiv:2306.11706},
  year={2023}
}

@article{huang2023embodied,
  title={An embodied generalist agent in 3d world},
  author={Huang, Jiangyong and Yong, Silong and Ma, Xiaojian and Linghu, Xiongkun and Li, Puhao and Wang, Yan and Li, Qing and Zhu, Song-Chun and Jia, Baoxiong and Huang, Siyuan},
  journal={arXiv preprint arXiv:2311.12871},
  year={2023}
}

@inproceedings{bharadhwaj2024roboagent,
  title={Roboagent: Generalization and efficiency in robot manipulation via semantic augmentations and action chunking},
  author={Bharadhwaj, Homanga and Vakil, Jay and Sharma, Mohit and Gupta, Abhinav and Tulsiani, Shubham and Kumar, Vikash},
  booktitle={2024 IEEE International Conference on Robotics and Automation (ICRA)},
  pages={4788--4795},
  year={2024},
  organization={IEEE}
}

@inproceedings{zhou2024navgpt,
  title={Navgpt: Explicit reasoning in vision-and-language navigation with large language models},
  author={Zhou, Gengze and Hong, Yicong and Wu, Qi},
  booktitle={Proceedings of the AAAI Conference on Artificial Intelligence},
  volume={38},
  number={7},
  pages={7641--7649},
  year={2024}
}

@article{wang2023voyager,
  title={Voyager: An open-ended embodied agent with large language models},
  author={Wang, Guanzhi and Xie, Yuqi and Jiang, Yunfan and Mandlekar, Ajay and Xiao, Chaowei and Zhu, Yuke and Fan, Linxi and Anandkumar, Anima},
  journal={arXiv preprint arXiv:2305.16291},
  year={2023}
}

@article{liu2024odyssey,
  title={Odyssey: Empowering minecraft agents with open-world skills},
  author={Liu, Shunyu and Li, Yaoru and Zhang, Kongcheng and Cui, Zhenyu and Fang, Wenkai and Zheng, Yuxuan and Zheng, Tongya and Song, Mingli},
  journal={arXiv preprint arXiv:2407.15325},
  year={2024}
}

@article{chen2025robogpt,
  title={Robogpt: an llm-based long-term decision-making embodied agent for instruction following tasks},
  author={Chen, Yaran and Cui, Wenbo and Chen, Yuanwen and Tan, Mining and Zhang, Xinyao and Liu, Jinrui and Li, Haoran and Zhao, Dongbin and Wang, He},
  journal={IEEE Transactions on Cognitive and Developmental Systems},
  year={2025},
  publisher={IEEE}
}

@article{liang2022code,
  title={Code as policies: Language model programs for embodied control},
  author={Liang, Jacky and Huang, Wenlong and Xia, Fei and Xu, Peng and Hausman, Karol and Ichter, Brian and Florence, Pete and Zeng, Andy},
  journal={arXiv preprint arXiv:2209.07753},
  year={2022}
}

@article{glocker2025llm,
  title={Llm-empowered embodied agent for memory-augmented task planning in household robotics},
  author={Glocker, Marc and H{\"o}nig, Peter and Hirschmanner, Matthias and Vincze, Markus},
  journal={arXiv preprint arXiv:2504.21716},
  year={2025}
}

@article{black2410pi0,
  title={$\pi$0: A vision-language-action flow model for general robot control. CoRR, abs/2410.24164, 2024. doi: 10.48550},
  author={Black, Kevin and Brown, Noah and Driess, Danny and Esmail, Adnan and others},
  journal={arXiv preprint ARXIV.2410.24164},
  year={2024}
}

@article{team2025gemini,
  title={Gemini robotics: Bringing ai into the physical world},
  author={Team, Gemini Robotics and Abeyruwan, Saminda and Ainslie, Joshua and Alayrac, Jean-Baptiste and Arenas, Montserrat Gonzalez and Armstrong, Travis and Balakrishna, Ashwin and Baruch, Robert and Bauza, Maria and Blokzijl, Michiel and others},
  journal={arXiv preprint arXiv:2503.20020},
  year={2025}
}

@inproceedings{wu2025momanipvla,
  title={Momanipvla: Transferring vision-language-action models for general mobile manipulation},
  author={Wu, Zhenyu and Zhou, Yuheng and Xu, Xiuwei and Wang, Ziwei and Yan, Haibin},
  booktitle={Proceedings of the Computer Vision and Pattern Recognition Conference},
  pages={1714--1723},
  year={2025}
}

@inproceedings{kannan2024smart,
  title={Smart-llm: Smart multi-agent robot task planning using large language models},
  author={Kannan, Shyam Sundar and Venkatesh, Vishnunandan LN and Min, Byung-Cheol},
  booktitle={2024 IEEE/RSJ International Conference on Intelligent Robots and Systems (IROS)},
  pages={12140--12147},
  year={2024},
  organization={IEEE}
}

@article{zhang2024lamma,
  title={Lamma-p: Generalizable multi-agent long-horizon task allocation and planning with lm-driven pddl planner},
  author={Zhang, Xiaopan and Qin, Hao and Wang, Fuquan and Dong, Yue and Li, Jiachen},
  journal={arXiv preprint arXiv:2409.20560},
  year={2024}
}

@article{duffy2000embodiment,
author = {Duffy, Brian and Joue, Gina},
year = {2000},
month = {01},
pages = {},
title = {Intelligent robots: The question of embodiment}
}

@article{guzman2022ios,
  author={Guzman, Luis and Morellas, Vassilios and Papanikolopoulos, Nikolaos},
  journal={IEEE Robotics and Automation Letters}, 
  title={Robotic Embodiment of Human-Like Motor Skills via Reinforcement Learning}, 
  year={2022},
  volume={7},
  number={2},
  pages={3711-3717},
  doi={10.1109/LRA.2022.3147453}}

@article{kim2024survey,
  title={A survey on integration of large language models with intelligent robots},
  author={Kim, Yeseung and Kim, Dohyun and Choi, Jieun and Park, Jisang and Oh, Nayoung and Park, Daehyung},
  journal={Intelligent Service Robotics},
  volume={17},
  number={5},
  pages={1091--1107},
  year={2024},
  publisher={Springer}
}

@article{urosa,
  title={Distributed AI Agents for Cognitive Underwater Robot Autonomy},
  author={Buchholz, Markus and Carlucho, Ignacio and Grimaldi, Michele and Petillot, Yvan R},
  journal={arXiv preprint arXiv:2507.23735},
  year={2025}
}

@article{bjorck2025gr00t,
  title={Gr00t n1: An open foundation model for generalist humanoid robots},
  author={Bjorck, Johan and Casta{\~n}eda, Fernando and Cherniadev, Nikita and Da, Xingye and Ding, Runyu and Fan, Linxi and Fang, Yu and Fox, Dieter and Hu, Fengyuan and Huang, Spencer and others},
  journal={arXiv preprint arXiv:2503.14734},
  year={2025}
}

@article{perlo2025embodied,
  title={Embodied AI: Emerging Risks and Opportunities for Policy Action},
  author={Perlo, Jared and Robey, Alexander and Barez, Fazl and Floridi, Luciano and M{\u{A}}{\'s}kander, Jakob},
  journal={arXiv preprint arXiv:2509.00117},
  year={2025}
}

@misc{hu2025mllmbasedvideoframe,
      title={M-LLM Based Video Frame Selection for Efficient Video Understanding}, 
      author={Kai Hu and Feng Gao and Xiaohan Nie and Peng Zhou and Son Tran and Tal Neiman and Lingyun Wang and Mubarak Shah and Raffay Hamid and Bing Yin and Trishul Chilimbi},
      year={2025},
      eprint={2502.19680},
      archivePrefix={arXiv},
      primaryClass={cs.CV},
      url={https://arxiv.org/abs/2502.19680}, 
}

@misc{lin2025captionskeyframeskeyscoremultimodal,
      title={From Captions to Keyframes: KeyScore for Multimodal Frame Scoring and Video-Language Understanding}, 
      author={Shih-Yao Lin and Sibendu Paul and Caren Chen},
      year={2025},
      eprint={2510.06509},
      archivePrefix={arXiv},
      primaryClass={cs.CV},
      url={https://arxiv.org/abs/2510.06509}, 
}

@techreport{Mueller2021AIActCost,
  author       = {Mueller, Benjamin},
  title        = {How Much Will the Artificial Intelligence Act Cost Europe?},
  institution  = {Center for Data Innovation, Information Technology and Innovation Foundation},
  year         = {2021},
  number       = {July 26, 2021},
  url          = {https://itif.org/publications/2021/07/26/how-much-will-artificial-intelligence-act-cost-europe/},
  note         = {Estimates the EU AI Act will cost Europe \texteuro{}31 billion over 5 years, reduce AI investment by ~20\%, impose up to \texteuro{}400,000 compliance costs on an SME deploying a high-risk AI system.}
}

@techreport{UKGov2023AIGovernanceImpact,
  author       = {{Department for Science, Innovation and Technology}},
  title        = {Evidence to Support the Analysis of Impacts for Artificial Intelligence Governance},
  institution  = {UK Government},
  year         = {2023},
  month        = {March},
  url          = {https://assets.publishing.service.gov.uk/media/64255c0460a35e000c0cb176/evidence_to_support_the_analysis_of_impacts_for_artifical_intelligence_governance.pdf},
  note         = {Evidence review underpinning the UK’s AI governance impact analysis.}
}

@article{explainable,
  title={Enhancing robot explanation capabilities through vision-language models: a preliminary study by interpreting visual inputs for improved human-robot interaction},
  author={Sobr{\'\i}n-Hidalgo, David and Gonz{\'a}lez-Santamarta, Miguel {\'A}ngel and Guerrero-Higueras, {\'A}ngel Manuel and Rodr{\'\i}guez-Lera, Francisco Javier and Matell{\'a}n-Olivera, Vicente},
  journal={arXiv preprint arXiv:2404.09705},
  year={2024}
}

@misc{llama_ros,
author       = {Miguel A. Gonzalez-Santamarta},
  howpublished = {\url{https://github.com/mgonzs13/llama_ros}}
}

@misc{stretch_ai,
author       = {HelloRobotInc},
  howpublished = {\url{https://github.com/hello-robot/stretch_ai}}
}

@article{feng2025multi,
  title={Multi-agent embodied ai: Advances and future directions},
  author={Feng, Zhaohan and Xue, Ruiqi and Yuan, Lei and Yu, Yang and Ding, Ning and Liu, Meiqin and Gao, Bingzhao and Sun, Jian and Zheng, Xinhu and Wang, Gang},
  journal={arXiv preprint arXiv:2505.05108},
  year={2025}
}

@article{buadicua2025contemporary,
  title={Contemporary Agent Technology: LLM-Driven Advancements vs Classic Multi-Agent Systems},
  author={B{\u{a}}dic{\u{a}}, Costin and B{\u{a}}dic{\u{a}}, Amelia and Ganzha, Maria and Ivanovi{\'c}, Mirjana and Paprzycki, Marcin and Seli{\c{s}}teanu, Dan and Wrona, Zofia},
  journal={arXiv preprint arXiv:2509.02515},
  year={2025}
}

@article{sapkota2505ai,
  title={AI Agents vs. Agentic AI: A Conceptual Taxonomy, Applications and Challenges. arXiv 2025},
  author={Sapkota, R and Roumeliotis, KI and Karkee, M},
  journal={arXiv preprint arXiv:2505.10468}
}

@online{anthropic2024building,
  author       = {Schluntz, Erik and Zhang, Barry},
  title        = {Building effective agents},
  year         = {2024},
  month        = dec # "~19",
  url          = {https://www.anthropic.com/engineering/building-effective-agents},
  note         = {Accessed: YYYY-MM-DD},
  organization = {Anthropic PBC}
}

@article{bandi2025rise,
  title={The Rise of Agentic AI: A Review of Definitions, Frameworks, Architectures, Applications, Evaluation Metrics, and Challenges},
  author={Bandi, Ajay and Kongari, Bhavani and Naguru, Roshini and Pasnoor, Sahitya and Vilipala, Sri Vidya},
  journal={Future Internet},
  volume={17},
  number={9},
  pages={404},
  year={2025},
  publisher={MDPI}
}

@article{raptis2025agentic,
  title={Agentic LLM-based robotic systems for real-world applications: a review on their agenticness and ethics},
  author={Raptis, Emmanuel K and Kapoutsis, Athanasios Ch and Kosmatopoulos, Elias B},
  journal={Frontiers in Robotics and AI},
  volume={12},
  pages={1605405},
  year={2025},
  publisher={Frontiers}
}
